# Theory of Machine Networks: A Case Study
by Rooz Mahdavian and Richard Diehl Martinez

## Abstract

We propose a simplification of the Theory-of-Mind Network architecture, which focuses on modeling complex, deterministic machines as a proxy for modeling nondeterministic, conscious entities. We then validate this architecture in the context of understanding engines, which, we argue, meet the required internal and external complexity to yield meaningful abstractions.

## Foundations

Our notion of theory-of-mind is drawn from [1], in which the authors propose that theory-of-mind is best encapsulated as the ability of individuals to impute mental states of **both** themselves and the people they interact with. These mental states are conditioned exclusively on an agent's previously observed behaviors, for the purpose of successfully predicted their future behaviors.

A neurological basis for this notion of theory-of-mind has been extensively studied. In particular, Firth et al. demonstrated that difficulties in acquiring theory-of-mind can be attributed to brain damage in the frontal cortex and in medial prefrontal region of the temporo-parietal junction. Previously, the medial prefrontal region had been known to activate when a subject was asked to reflect upon their own mental states. Similarly, the temporo-parietal junction was particularly stimulated in subjects when asked to reason about perceptual cues to recognize the actions and intentions of other agents. Theory-of-mind can thus be understood as a tool that enables humans to bootstrap better predictions about an agent's behavior by modeling their internal mental states. Theory-of-mind consequent significance, then, lies in understanding human desire and intentions.

A number of computational approaches have attempted to model this biological property, and we focus on the recent proposal of a Theory of Mind Networks (ToMnet) framework by Rabinowitz et al [2]. The authors argue that predictions about an agent's future behavior can be derived from learned mental and character embeddings of these agents. In their proposed architecture, the character embedding encodes a history of an agents' past behaviors over a series of time steps. The mental embedding is similarly derived by encoding the current interaction episode up to the current period, conditioned on a fixed character embedding. By leveraging these two models, Rabinowitz et al. demonstrate that observers can learn to build effective prior distributions which encode the commonalities between agents in the population. These in turn allow the model to more rapidly and effectively build predictions about an agent's future behavior.

## Introduction: Minds and Machines

The fundamental goal of a Theory-of-Mind Network (biological or artificial) is in learning a representation of an entity which is useful for predicting the subsequent behavior of that entity. To this end, the ToMnet architecture learns representations of an agent's character and mental state to estimate the probabilities governing an agent's policies. In other words, it targets non-deterministic agents.

This is, of course, a perfectly valid approach to modeling human behavior, as it's an open question as to whether or not (at the lowest levels) our behavior is deterministic. But it also presents a problem in validating the accuracy of this approach: the POMDP distributions it estimates for a given agent must be known, and this is not necessarily the case for human agents.

The ToMnet authors themselves note that a Theory-of-Mind model "should make little to no reference to the agents' true underlying structure", since, in the context of human theory-of-mind, "we do not typically attempt to estimate the activity of others' neurons, infer the connectivity of their prefrontal cortices, or plan interactions with a detailed approximation of the dynamics of others hippocampal maps". In this, the authors imply the possibility that a representation of the *mechanical* structure of another's brain, as complex as it would be, could lead to very strong predictions about their behavior; put simply, that the mind in theory-of-mind operates as a *machine*. With this possibility in mind, we choose to focus on learning to represent entities which are highly complex, but are still fundamentally deterministic, as a proxy for learning to represent systems which could then theoretically scale to the complexity of a deterministic mind. The resulting architecture is, of course, very similar, and it's goals are fundamentally the same.

## Complexity

Importantly, a Theory-of-Machine Network retains the primary noun; that is, the goal is still to develop an abstract representation, or *theory*, of a machine, rather than to directly learn the underlying machinery. To be a interesting representation then, the interactions between the underlying machinery should be sufficiently high-dimensional, for which a direct mapping would be inordinately complex and likely computationally infeasible. But importantly, by our assumption of determinism, we then know that a direct mapping *does* in-fact exist, which makes the problem of representing it abstractly tractable.

Further, the machine's complexity should yield states and their emergent properties; that is, the internal mechanisms of the machine should interact through time to yield different configurations, which in-turn produces different observable behavior, and thus this notion of state should consequently be necessary in understanding this machines behavior. Or, the machines deterministic behavior should be a product of *both* its fixed properties and its dynamic state.

Any machine which doesn't meet the above criterion is likely too simple to represent with a meaningful abstraction, as opposed to a direct mapping (low-dimensional but stateful machine) or dimensionality reduction (high-dimensional but state-less machine). With this complexity in mind, we now focus on designing an architecture which yields useful abstractions.

## Architecture

Akin to ToMnets, Theory-of-Machine Networks measure the quality of a learned abstraction purely in its direct usefulness in predicting a machine's behavior, and frame the task as meta-learning, where the abstraction for a particular machine is learned to *be learned* from real-time observations, rather than learned directly. But the nature of this meta-learning is our most significant departure from Theory-of-Mind Networks: since we are focused on deterministic *machines*, we now observe **I/O** - pairs of inputs the machine and the corresponding outputs - rather than POMDP trajectories.

With the complexity constraints previously considered, I/O pairs should be observed in sequence, such that the Theory-of-Machine Network learns how to encode **both** the *underlying mechanics* of the machine through isolated I/O, and the *current state* of the machine by observing patterns in I/O through time. These in-sequence observations yield the **stateful machine-embedding**:

$$\vec{s}_t = f_\theta\big(\{I_i^{(obs)}, O_i^{(obs)}\}_{i=(t-n)}^{t}\big)$$

where $t$ is the current time-step, $n$ is the sequence length, and $f_\theta$ is a learned network. As we continue to observe a particular machine in progressive sequences, we continue to update the stateful machine-embedding, to yield an abstraction which captures the short-term state of the machine, and it's underlying machinery in that state. But the usefulness of this embedding is then highly dependent on a fixed sequence length, when ideally our understanding should monodically improve with more observations.

More directly, the stateful machine embedding defines a posterior on the machine in it's current state, but the understanding of machine is then missing a prior.

The Theory-of-Mind Network solves this by introducing a separate encoder for previously observed POMDP trajectories, but we'd like our model to work entirely in real-time. So, we derive a **stateless machine-embedding** as the decayed sum of stateful machine-embeddings through time:

$$\overrightarrow{m}_t = \beta \circ \overrightarrow{m}_t + \gamma \circ \vec{s}_t$$

where $\beta$ and $\gamma$ are learned parameters of the same dimensionality as our stateful and stateless machine embeddings. Together, $\vec{s}_t$ and $\overrightarrow{m}_t$ are then fed-forward alongside $I^{(obs)}_{(t+1)}$ to a linear **theory network**, which integrates over these embeddings and the given input to predict the subsequent output, $O^{(obs)}_{(t+1)}$. This network can then be trained entirely end-to-end (in both online and offline environments); a visualization is provided below.

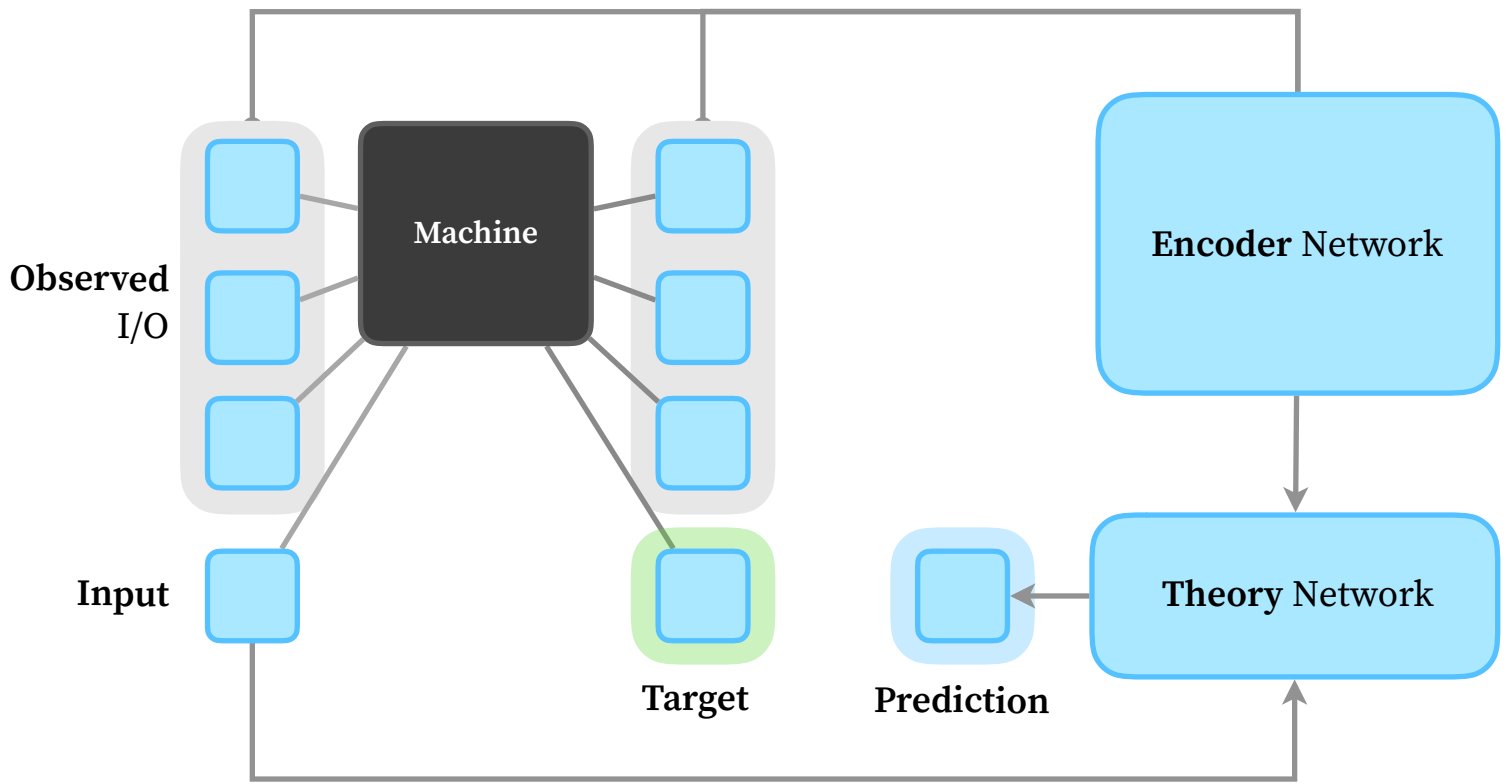

**Figure 1: Theory-of-Machine Network Architecture**
Pairwise mappings of observed I/O in sequence are encoded into a stateful machine-embedding, which is then used to predict machine output at the subsequent time-step, given the input at that time-step. The stateless machine-embedding is hidden for simplicity.

We now focus our attention on validating this architecture with a machine-specific case study.

## Case Study: Engines

**Motivation**
Per the complexity constraints outlined previously, we require a machine with internal mechanics which exist in a very high-dimensional space, and with emergent states as these mechanics interact through time. A CPU would be one such machine. A circuit diagram of the human brain would be another. But these representing either of these machines abstractly is still far too daunting a task for a baseline result. So, we turn to a machine of meaningful complexity, but with an abstraction we think is still achievable: the engine of a car.

Note that by *engine*, we are actually referring to all of the internal mechanics which translate an input control (the deltas in throttle, braking, and steering) to output movement (the positional deltas in $x$, $y$, and $z$ coordinates) - the physical engine, of course, plays a role, but so does the differential, the transmission, the anti-braking system, the traction control system, and so forth. The engine, in this sense, is also parameterized by a number of features: the mass of the vehicle, its center of gravity, the grip of tyres, the grip of the brakes, and much more, all of which can result in meaningful changes in output. Finally, the engine exists in **both** internal and external states. Internally, states are defined by the temperature of the physical engine, the temperature of the tyres, the wear in the tyres, the current oiling of the gears, and so forth. Externally, states are defined by the grip of the road, the incline of the road, the positional velocity of the car, the rotational velocity of the car, and more. Altogether, the stateful, high-dimensional space in which this machine resides makes learning to abstract meaningful representations of a particular engine purely from I/O observations a significant task. But crucially, there necessarily exists a function which, given each one of these latent variables, maps precisely from input to output, as it is simply a nontrivial integration over Newtonian mechanics.

**Simulator**
To begin, we needed an environment to observe engine machines within, to create our training loop. Fortunately, a number of sophisticated simulation engines exist, which perform the exact sort of Newtonian integration described earlier, over *every* latent variable mentioned (and many more). We choose to target *Assetto Corsa* [10], as it offers a Python API to read I/O values from an engine in the simulator, and also provides an agent which provides inputs to the engines to drive a car around a given track. And importantly, the underlying physics engine exposed to the agent-piloted car is identical to that exposed to a player-piloted car, which means we can synthesize a large, meaningful dataset of observed I/O.

**Dataset**

We then synthesized a dataset of over 60,000 labeled I/O pairs, sampling from *Assetto* at 10HZ via RPC. The track for the agent-driven policy was fixed to Nürburgring, a benchmarking track in competitive motorsport famous for its strain on a car's engine, due to a number of sudden shifts in grade, lane width, and turn radius. Ideally, this would result in a strong variance in engine I/O. The 54 cars then run on the track were selected for largest possible extra-class variance (SUVs, hatchbacks, F1s, GTs, supers, and more) while retaining some cars with little intra-class variance (to sanity check the corresponding representations).

The resulting dataset was then split into a train and test pair, where 42 cars were sampled for the training set, and 12 cars were sampled for the test set. Crucially, the test set is then composed of *engines* which the network hasn't seen before, as opposed to novel observed I/O from a car it has previously observed.

**Performance**

We then implemented a Theory-of-Machine Network in PyTorch, and trained it on Google Cloud with a sequence length of 100.

The loss was measured as the mean-squared-error between the predicted $(dx, dy, dz)$ and the labeled values for a given sequence. Within 30 epochs, the network achieved an MSE loss of 0.004, which is near-perfect for the scale of our data (the range of each target value is between ~0.1 and 60). The corresponding loss on the test set was 0.008, which suggested incredibly strong generalization and performance in general. As the engine representations resulted in near-perfect accuracy in prediction of engine behavior, we then focused on qualitatively evaluating the resulting abstractions.

**Visualizations**

We computed stateful machine embedding for and collected metadata from *Assetto* on each of the cars in collective training and test set, and performed PCA to project the embeddings into 3-dimensions, which were then colored with different meta-tags for qualitative evaluation. These are visualized below.

Of note is that since each embedding is a stateful machine embedding, meaning that it was computed from a random sample of 100 I/O pairs from that engine's respective dataset. Thus, the stretch of observed track is not held consistent across the embeddings, meaning the variance observed by the model is not held consistent in the embeddings it produces (as the way it might encode a particular engine is thus substantially different if it observes it in a straight or a turn). Further, the engine machine is, as discussed, a complex, balanced integration over many latent variables; thus, colorization by any individual variable alone is not very likely to yield consistent clusters. Still, the visualizations have interesting patterns, further suggesting that the network has, indeed, learned to produce meaningful abstractions.

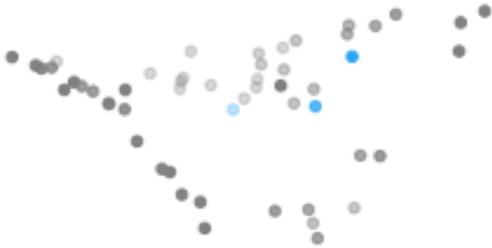

**Figure 2: Engine Type, SUV**
SUVs are, as expected clustered tightly. This is more than likely due to their extremely high masses, which, at those extremes, would likely alone have a much larger influence on engine I/O. This is cross-validated in Figure 7.

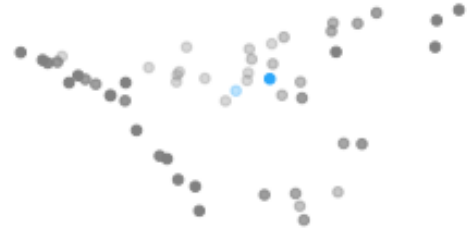

**Figure 3: Engine Type, Track**
Track engines also exhibit strong clustering. This, conversely, is likely due to their extremely low masses, for reasons akin to SUVs.

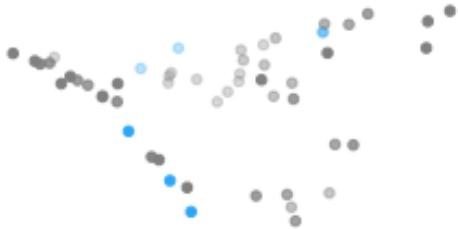

**Figure 4: Engine Type, F1**
F1 engines exhibit consistent clustering along a particular fixed. This is likely due to an axis of generalities (higher torque, lower mass) of which there are intra-car variations.

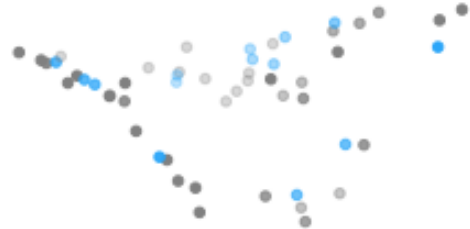

**Figure 5: Engine Type, Sport**
Sportscars, as expected, exhibit the highest-variation and least-consistent clustering. This is likely due to how broad a class it is; unlike the previous three, it does not imply any particular extremes in the latent variables.

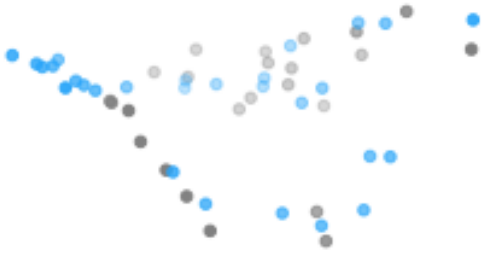 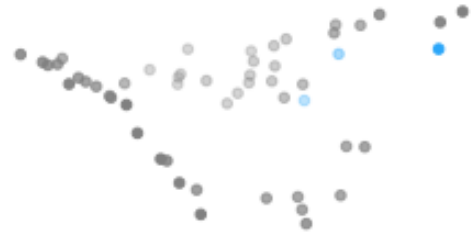

**Figure 6: Weight, 1000-1500kg**
Similar to Figure 5, these are median values for weight, and thus its likely within this range weight alone is not a strong factor in engine I/O.

**Figure 7: Weight, 2000-2500kg**
As expected, more extremes values in weight result in much tighter clustering, suggesting that the extremes of weight play a more dominant role in engine I/O.

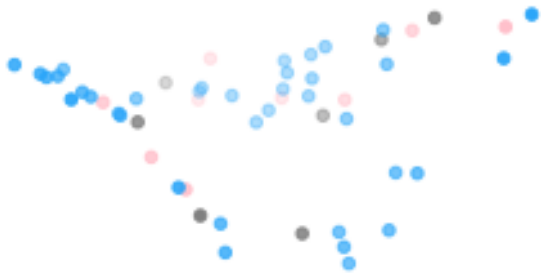

**Figure 8: Year, 1960-2020**
The different 20-year ranges of engine manufacturing produce, as they should, no discernible pattern. The year alone should tell you almost nothing about engine I/O.

## Conclusion

We have proposed a simplified variant of Theory-of-Mind networks which focuses on abstracting a machine purely from sequential observations of I/O, and performed a baseline validation of our architecture in abstracting engines, with incredibly promising initial results.

Although the test accuracy was near-perfect, significantly more work needs to be done to interpret the quality and depth of the meaning captured in the computed representations. As the network learns to produce representations which are useful in predicting subsequent machine behavior, it follows that the engine machine representations should be useful in a number of interesting tasks. You should, for example, be able to train a simple, linear network which maps one engine's optimal driving policy to another engine's optimal policy, given their respective engine representations. Or, train an RL algorithm which learns the optimal sequential variations in engine I/O to maximize the variance in the computed engine representation - in effect, learning how to quickly learn the unique properties of a particular engine. The combination of these two models could then, hypothetically, generalize a fixed driving policy to the mechanical limits of a wide number of different cars, after driving a particular unseen car for a very short amount of time (as a professional racer can do today). All of these is necessary work in understand the richness of the information being encoded by the Theory-of-Machine network.

Fundamentally, we believe successful architectures which abstract extremely intricate machines through observations of there I/O could scale well to increasingly complex systems (given more data), and that eventually, rich abstractions of a human brain purely from its I/O may be within reach. This, of course, depends entirely on our minds being machines.

And at this point, really, it's anyone's guess.